\documentclass[runningheads]{llncs}

\usepackage{microtype}
\usepackage{graphicx}
\usepackage{subfigure}
\usepackage{booktabs} 
\usepackage{hyperref}

\titlerunning{Gator: Customizable Channel Pruning of Neural Networks with Gating}

\usepackage{euscript}
\usepackage{amsmath}
\usepackage[flushleft]{threeparttable}
\usepackage{graphicx}
\usepackage{etoolbox}
\appto\TPTnoteSettings{\footnotesize}
\usepackage{hyperref}
\usepackage{textpos}

\title{Gator: Customizable Channel Pruning \\
of Neural Networks with Gating }

\author{Eli Passov\inst{} \and Eli O.~David\inst{} \and Nathan S. Netanyahu\inst{}\thanks{Nathan Netanyahu is also affiliated with the Department of Computer Science at the College of Law and Business, Ramat-Gan 5257346, Israel} }

\authorrunning{E.~Passov, E.O.~David and N.S.~Netanyahu}

\institute{
Department of Computer Science, Bar-Ilan University, Ramat-Gan 5290002, Israel \\
\email{elipassov@gmail.com, mail@elidavid.com, nathan@cs.biu.ac.il}
}

\makeatletter
\def\thickhline{%
  \noalign{\ifnum0=`}\fi\hrule \@height \thickarrayrulewidth \futurelet
   \reserved@a\@xthickhline}
\def\@xthickhline{\ifx\reserved@a\thickhline
               \vskip\doublerulesep
               \vskip-\thickarrayrulewidth
             \fi
      \ifnum0=`{\fi}}
\makeatother

\newlength{\thickarrayrulewidth}
\newcommand\clearrow{\global\let\rowmac\relax}
\clearrow

\begin{document}

\maketitle

\begin{textblock*}{5in}(0mm, -65mm)
{\textbf{Ref:} An earlier version of this arXiv appeared at the
\textit{International Conference
on Artificial Neural Networks (ICANN)}, Springer LNCS, Vol.~12894, pp.~46--58,
Bratislava, Slovakia, September 2021.}
\end{textblock*}

\begin{abstract}
The rise of neural network (NN) applications
has prompted an increased interest in compression, with a particular focus on \textit{channel pruning}, which does not require any additional hardware. 
Most pruning methods employ either single-layer operations or global schemes to determine which channels to remove followed by fine-tuning of the network. 
In this paper we present Gator, a channel-pruning method which temporarily adds learned gating mechanisms for pruning of individual channels, and which is trained with an additional auxiliary loss, aimed at reducing the computational cost due to memory, (theoretical) speedup (in terms of FLOPs), and practical, hardware-specific speedup.
Gator introduces a new formulation of dependencies between NN layers which, in contrast to most previous methods, enables pruning of non-sequential parts, such as layers on ResNet's highway, and even removing entire ResNet blocks.
Gator's pruning for ResNet-50 trained on ImageNet produces state-of-the-art (SOTA) results, such as $50\%$ FLOPs reduction with only $0.4\%$-drop in top-5 accuracy. 
Also, Gator outperforms previous pruning models, in terms of GPU latency by running 1.4 times faster. 
Furthermore, Gator achieves improved top-5 accuracy results, compared to MobileNetV2 and SqueezeNet, for similar runtimes. 
The source code of this work is available at: \url{https://github.com/EliPassov/gator}.
\end{abstract}

\section{Introduction}

The use of NNs, in particular, \textit{convolutional neural networks} (CNNs), in a variety of Computer Vision applications has surged considerably in recent years. 
Most NNs consist of tens to hundreds of millions of weight parameters and require typically billions of \textit{floating point operations} (FLOPs) for processing a single image. This limits their use on available hardware, especially on small devices. Consequently, many works in recent years have focused on reducing the computational cost of network architectures without severely impairing their performance, either by creating more efficient network architectures or by applying network compression. 
Pruning is a popular method among compression methods, where network weights and activations are removed. 
Pruning methods can be categorized as \textit{unstructured}, i.e., referring to the removal of individual weights, or  \textit{structured}, i.e., referring mostly to the removal of convolutional channels. Unstructured pruning is considered much less practical, as it requires special hardware for efficient application.

More recent techniques also utilize dynamic computation in conjunction with \textit{gating} mechanisms to control
the network operations that are invoked for each input instance; thus, the computational cost is determined dynamically for each individual input. Our method utilizes dynamic gating mechanisms only during training, to learn which channels should be removed, thereby 
producing a pruned network without any gating mechanisms.

The effectiveness of pruning methods is usually assessed by these parameters:
\begin{itemize}
\item
Reduction of memory required for storing weights;
\item
Theoretical speedup, measured by the reduction of the number of FLOPs;
\item
Latency/runtime on a specific hardware.
\end{itemize}
Note that latency is not commonly used as an optimization objective, and that theoretical speedup is never translated into the exact same practical speedup. For unstructured methods, this is due to the fact that most hardware does not support sparse operations, leading to no optimization in practice; for structured pruning, the sole focus on the FLOPs metric ignores hardware-specific behavior.

This paper makes the following contributions:
\begin{itemize}
\item Presentation of \textit{Gator}, an innovative pruning technique that utilizes channel hard-gating
to remove channels during training while also adjusting the remaining weights;
\item Introduction of a novel formulation for dependencies between NN layers allowing Gator to prune non-sequential parts of NN architectures (such as ResNet's highway), which are not handled by most pruning methods;
\item Presentation of a customizable cost function, which
reflects various practical compression criteria, such as memory reduction, FLOPs reduction, and hardware-specific speedup;
\item Obtaining new SOTA results for ResNet-50 pruning on ImageNet, in particular, 50\% FLOPs reduction with $0.4\%$ top-5 accuracy drop and runtime speedup (latency reduction) of 1.4 for the same accuracy as other methods; 
\item Presentation of additional results for higher pruning rates, outperforming MobileNet V2 and SqueezeNet for the same runtime.

\end{itemize}

\section{Related Work}

\subsection{Network Compression}

Most methods for NN compression can be categorized into three categories. First, tensor decomposition (\cite{lebedev2014speeding}, \cite{jaderberg2014speeding}, \cite{kim2015compression}), where the convolutional operation is approximated by decomposing it to smaller operations. 

Second, quantization methods (\cite{zhou2017incremental}, \cite{zhu2016trained}, \cite{chen2015compressing}), in particular, binary methods (\cite{rastegari2016xnor}, \cite{courbariaux2016binarized}), where the network weights are approximated with lower precision variables allowing for memory reduction.

And third, pruning, where parts of layer outputs or weights are removed. In early works \cite{lecun1989optimal} and more recent ones \cite{han2015learning}, pruning methods are characterized by weight elimination and a "fine tuning" training stage. A thorough review of pruning methods can be found in \cite{blalock2020state}, which also highlights the distinction between structured and unstructured pruning. Since unstructured pruning cannot reduce computational costs on most hardware types, we will focus and compare Gator only to other structured pruning methods.

\subsection{Structured Channel Pruning}

Most channel pruning works differ in the criteria by which channels are pruned. In \cite{li2016pruning} the smallest sums of weights of channels from all layers are pruned, while non sequential layers are being considered. 
The pruning criterion in~\cite{molchanov2016pruning} is the first order Taylor expansion of the loss ($w\frac{\partial L}{\partial w}$).
Weights closest to the geometric median are pruned in \cite{he2019filter}, in an iterative process for each layer. 
And in~\cite{yamamoto2018pcas}, individually trained squeeze and excitation \cite{hu2018squeeze} blocks are used to prune layers with the lower attention values.

Other methods, such as \cite{he2017channel}, remove the convolutional input channels with the least impact on the output.  
While~\cite{zhuang2018discrimination} first train auxiliary classification losses on intermediate layers and define reconstruction losses on potential channels to be remove. The channels chosen to be removed are those which have the smallest impact on the sum total of all classification and reconstruction losses.

\subsection{Gating}

Dynamic gating modules determine, for each individual input, which parts are executed and which are skipped within a computational NN graph. 
In~\cite{veit2018convolutional}, squeeze and excitation blocks \cite{hu2018squeeze} are used with a gumbel softmax \cite{jang2016categorical} gate on individual residual blocks, enabling skipping some blocks dynamically. In~\cite{ehteshami2019batch}, the same gating mechanism is applied on individual channels.

Methods which do not rely on gumbel softmax, such as \cite{gao2018dynamic}, use regression on the global pooling vector of the convolutional input to determine dynamically which channels are passed to the convolution. In \cite{hua2019channel}, convolutions are divided into two parts, such that the output of the first gates the execution of the second.

A soft sigmoid gate is used in \cite{gain2020adaptive} to prune connections in fully-connected layers. Our method also uses a sigmoid; however, it gates entire channels, using a hard gating mechanism with a different learning process which leads to practical results on ImageNet.

\subsection{Network Architecture Search}

Network architecture search (NAS) first introduced in~\cite{zoph2016neural}. It focuses on finding an optimal network architecture within a wide search space using methods such as reinforcement learning. Many NAS works, such as \cite{tan2019efficientnet}, typically regard the number of layer channels as part of the search space while minimizing the FLOPs metric. The search space of \cite{he2018amc} consists of layer pruning ratios minimizing either FLOPs or memory metrics for a given target accuracy. Other methods, such as \cite{wu2019fbnet}, optimize the latency metric. Although many NAS methods also share the idea of optimizing the number FLOPs, memory footprint or latency, and even yield SOTA performance in many cases, they are extremely computationally intensive compared to pruning methods, and thus exhibit sub-optimal latency.

\section{Approach}

In this section we describe the components of Gator. We start with the gating mechanism in Subsection~\ref{channel_gating}, and then explain in Subsection~\ref{hypergraph} how modeling the network architecture as a hypergraph encapsulates the channel dependency between different layers in the network architecture, thereby allowing for non-sequential pruning. Finally, in Subsection~\ref{computational_cost} we present the loss function for computational cost and demonstrate how to customize it for FLOPs, memory, and latency reduction.

\subsection{Channel Gating} \label{channel_gating}

\begin{figure}
\vskip -0.3in
\begin{center}
\centerline{\includegraphics[width=10cm]{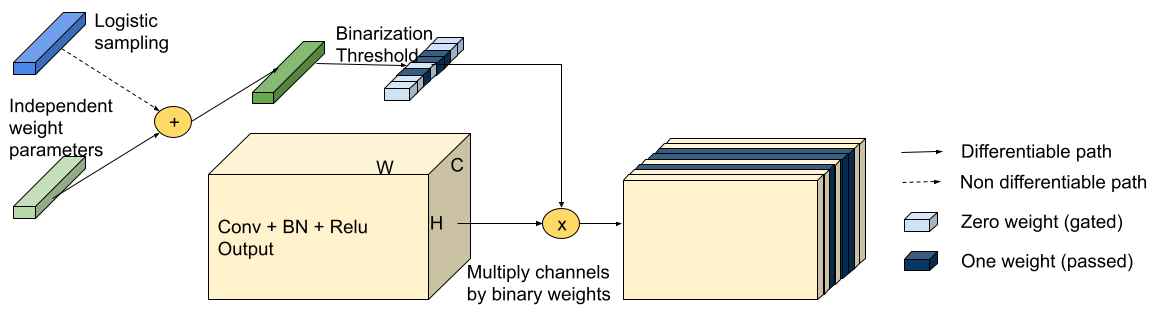}}
\caption{Logistic sigmoid gating model for an output of a layer.}
\label{fig:gating_mechanism}
\end{center}
\vskip -0.5in
\end{figure}

Gator utilizes a gating mechanism which determines for each individual channel whether it should be passed or gated for each individual sample within a batch. 
The gating mechanism (see Eqn.~\ref{eq:logistic_sigmoid})
uses a binary version of the Gumbel softmax distribution (introduced in~\cite{jang2016categorical}), referred to as the \textit{logistic sigmoid distribution}.
\begin{equation}
\label{eq:logistic_sigmoid}
\begin{split}
& \sigma(\theta) = \frac{1}{1+e^{-\frac{\theta+x}{\tau}}} \quad
{\rm where} \quad x\sim Logistic(0, 1) \\
& g(\theta) = \begin{cases}
1 & \sigma(\frac{\theta+x}{\tau}) \geq 0.5 \\
0 & \sigma(\frac{\theta+x}{\tau}) < 0.5 \\
\end{cases} \simeq sign(\theta+x)\; .
\end{split}
\end{equation}
The gating binary value $g(\theta)$ is determined by adding the gating weight $\theta$ to a random variable $x$ sampled from a logistic distribution. The purpose of the sigmoid function is to enable a continuous gradient while the threshold test provides a discrete output equivalent to a sign function in the forward pass. 
The temperature parameter $\tau$, which controls the gradient magnitude, is set to the default value of 1 and is, henceforth, ignored. 
The addition of the random variable $x$ is according to the reparameterization trick in~\cite{kingma2013auto},
which enables a computational flow of the gradients for the gating weights
as part of the backpropagation process,
while providing a discrete value in the forward pass.

A gating module can be placed before or after a layer to pass or block individual channels, thereby simulating the effect of pruning input or output channels. Fig.~\ref{fig:gating_mechanism} illustrates gating on an output of a model.

\subsection{Hypergraph of Neural Network Channels}\label{hypergraph}

\begin{figure}[ht]
\begin{center}
\centerline{\includegraphics[width=\columnwidth]{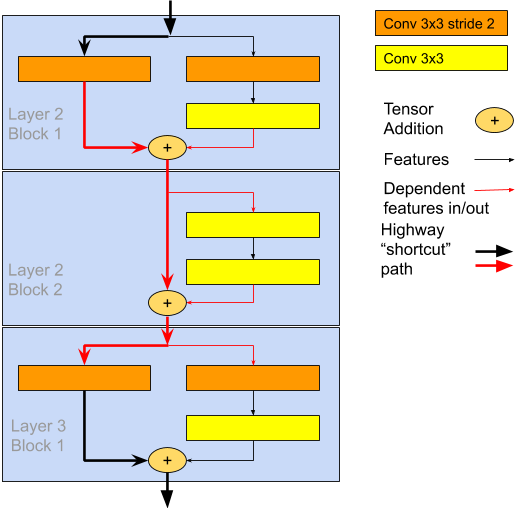}}
\caption{Layer 2 blocks 1,2 and layer 3 block 1 from ResNet-18 architecture, demonstrating channel dependency between 6 different convolutions illustrated by red lines; Thicker lines indicate the highway path which is the main cause of this dependency.}
\label{fig:resnet_connections}
\end{center}
\vskip -0.4in
\end{figure}

Early pruning methods on sequential network architectures, such as VGG~\cite{simonyan2014very}, focus naturally on individual convolutions, so they are applicable to all layers. 
However, most recent architectures consist of many non-sequential parts. 
Considering ResNet architectures, for example, note that inputs of some convolutions depend on outputs of multiple convolutions residing in different ResNet blocks, as a result of the ``highway'' path configuration. 
Thus, using a channel pruning scheme, in an attempt to remove a channel from the output of a single convolution may not remove the same channel from the input of a dependent convolution or the output of another convolution, as implied by the network's non-sequential architecture. In other words, one needs to consider the impact of a channel removal on the input and output channels of \textit{all} convolutions dependent on that channel. 
This aspect is illustrated for the ResNet-18 architecture in Fig.\ref{fig:resnet_connections}.
We shall refer to a collection of channel-dependent convolutions, as far as their input and output channels are concerned, as a \textit{channel dependency group}.

We define an undirectional hypergraph $H=(V,E)$, where the set of vertices $V$ corresponds to the union of the input and output channel sets over all convolutions. Let $C^{in}$ and $C^{out}$ denote these input and output sets, respectively. Specifically, the vertices $c_i^{in} \in C^{in}$ and $c_i^{out} \in C^{out}$ represent the (numbers of the) input and output channels of convolution $i$, respectively.
The set of edges $E$ is composed of subsets of $V$, where an edge $e_j \in E$ corresponds to channel dependency group $j$. (Each edge in $E$ is referred to as a \textit{dependency edge}.) We represent the dependency group as $e_j = C_j^{in} \cup C_j^{out} \subseteq V$, where $C_j^{in} \subseteq C^{in}$ and $C_j^{out} \subseteq C^{out}$ denote, respectively, the input and output channel subsets associated with dependency group $j$.
We can now define pruning as a removal of channels from each dependency edge.

For sequential networks, each dependency edge is composed of a pair of a convolutional layer's output and the following convolutional layer's input.
However, we can now also map channel dependency for non-sequential networks, such as ResNet-50, as detailed in Table~\ref{tbl:resnet_mapping}.
Specifically, 53 convolutions and a fully-connected layer are mapped to 37 dependency edges, 32 of which are trivial edges representing sequential connections inside bottleneck blocks, and to 5 additional edges, which include more than two convolutions (some of which are located along the ``highway'' path).

\begin{table}[!tbp]
\caption{Dependency mapping for ResNet-50; Each row represents one edge composed of output and input layers (vertices); encoding used, L1B1-(example of) layer and block, C0-first convolution, C1,C2,C3-bottleneck convolutions, D-downsampling convolution, and FC-fully connected; for brevity, trivial edges inside bottlenecks are listed once}
\label{tbl:resnet_mapping}
\vskip -0.2in
\begin{center}
\begin{small}
\begin{sc}
\begin{tabular}{|c|c|c|}
\hline
output layers & input layers & \# edges \\ [0.5ex] 
\hline
c0 & l1b1d, l1b1c1 & 1 \\ \hline
c1 & c2 & 16\\ \hline
c2 & c3 & 16\\ \hline
l1b1d, l1b1c3, l1b2c3, l1b3c3  & 
l1b2c1, l1b3c1, l2b1d, l2b1c1  & 1 \\ \hline
\begin{tabular}{@{}c@{}} l2b1d, l2b1c3, l2b2c3, l2b3c3, \\ l2b4c3 \end{tabular}&
\begin{tabular}{@{}c@{}} l2b2c1, l2b3c1, l2b4c1, l3b1d, \\ l3b1c1 \end{tabular} & 1 \\ \hline
\begin{tabular}{@{}c@{}} l3b1d, l3b1c3, l3b2c3 l3b3c3, \\ l3b4c3, l3b5c3, l3b6c3 \end{tabular} &
\begin{tabular}{@{}c@{}} l3b2c1, l3b3c1, l3b4c1, l3b5c1, \\ l3b6c1, l4b1d, l4b1c1 \end{tabular} & 1 \\ \hline
l4b1d, l4b1c3, l4b2c3, l4b3c3  & l4b2c1, l4b3c1, fc & 1 \\
\hline
\end{tabular}
\end{sc}
\end{small}
\end{center}
\vskip -0.3in
\end{table}

\subsection{Loss Function for Computational Cost} \label{computational_cost}

For each edge $e_j$ we have $c_j(t)$ non-gated channels at iteration $t$ from a total of $c_j$ channels, computed from the binary gating values defined in Eqn. (\ref{eq:logistic_sigmoid}):
\begin{equation}
\label{eq:channel_sum}
c_j(t)=\sum\limits_{i=1}^{c_j}g(\theta_j^i)\; .
\end{equation}
Let $\lambda_j(t)$ denote the per-channel cost factor for iteration $t$. The change in cost for different iterations stems from the fact that the number of non-gated input and output channels of a convolution changes at each iteration.

We now define the auxiliary computational loss 
\begin{equation}
\label{eq:computation_loss}
L_{computation}(t) = \sum\limits_{j=1}^m c_j(t) \widehat{\lambda_j}(t) = \frac{\sum\limits_{j=1}^m c_j(t)\lambda_j(t)}{\sum\limits_{j=1}^m c_j \lambda_j} \; ,
\end{equation}
where $m$ is the number of dependency edges, $\widehat{\lambda_j}(t)$ is the normalized, per-channel cost factor, and $\lambda_j$ is the initial cost factor.
The normalization sets the initial loss to $L_{computation}(0)=1$, so that it is easier to calibrate the total loss
\begin{equation}
\label{eq:total_loss}
L_{total}(t)=L_{original} + \alpha L_{computation}(t) \; ,
\end{equation}
 where $L_{original}$ is the original loss and $\alpha$ is a global weight being calibrated.

The memory footprint cost factor is the number of weights in all layers of each individual edge, i.e.,
\begin{equation} \label{eq:mem_footprint}
\lambda_j(t) = \sum\limits_{q_i \in \bf{C}_j^{in}} k_{q_i}^w k_{q_i}^h c_{q_i}^{out}(t) + \sum\limits_{q_i \in \bf{C}_j^{out}} k_{q_i}^w k_{q_i}^h c_{q_i}^{in}(t)\; ,
\end{equation}
where $\bf{C}_j^{out}$ and $\bf{C}_j^{in}$
are the convolution sets corresponding to $C_j^{out}$ and $C_j^{in}$, respectively.
For each convolution $q_i$, $k_{q_i}^w$ and $k_{q_i}^h$ denote the kernel width and kernel height, respectively. Also, $c_{q_i}^{in}(t)$ and $c_{q_i}^{out}(t)$ denote, respectively, the number of non-gated input and output channels at iteration $t$. 

The theoretical speed cost factor
is computed for a single input pixel, by multiplying the number of weights of each layer by the downsample factor $d_i$, i.e., the ratio between the number of pixels in the current layer's output and the original input image. Typically $d_i=4^{-s_i}$, where $s_i$ is the number of stride-2 operations (e.g., convolutions, maxpooling, etc.) 
encountered on a computational path from the input image to the current layer $i$. Thus the FLOPs cost factor is given by
\begin{equation} \label{eq:flop_cost}
\lambda_j(t) = \sum\limits_{q_i \in \bf{C}_j^{in}} d_i k_{q_i}^w k_{q_i}^h q_i^{out}(t) + \sum\limits_{q_i \in \bf{C}_j^{out}} d_i k_{q_i}^w k_{q_i}^h q_i^{in}(t)\; .
\end{equation}
 
For latency, the cost factor cannot be generally defined, as it is hardware-specific. Instead, we approximate, in this case, the cost factor $\lambda_j(t)$, by measuring the latency difference between the original network and a network with $50\%$ of the channels in $e_j$ pruned.

\subsection{Learning Rate Gradient Update Adjustment}\label{gradient_adjustment}

During training, the loss function described in Eqn.~\ref{eq:computation_loss}, produces a gradient update equivalent to the cost factor. The cost factor can be quite different in scale for different layer. For FLOPs cost, for instance, the cost factor is smaller for deeper layers (since they have more channels). As a result, pruning of deeper layers will be slower (take more iterations) compared to shallow layers. In turn, this could also lead to over pruning of shallow layers, causing extensive adjustment of the rest of the network weights to compensate, which eventually produces a sub optimal pruning result.

To combat both issues, we define a different time-dependent learning rate for the weights of each individual dependency edge 
\begin{equation} \label{eq:gradient_adjustment}
\eta_j(t) = \frac{\gamma \eta(t)}{\widehat{\lambda_j}(t)}\; ,
\end{equation}
where $\eta_j(t)$ is the learning rate of dependency edge $j$, $\eta(t)$ is the global learning rate, and $\gamma$ is a constant parameter. 

Breaking down the gradient update for an individual gating weight $i$ on dependency edge $j$, $\theta_j^i$, with a global pruning weight $\alpha$ we get
\begin{equation} \label{eq:gradient_adjustment_derivation}
\begin{split}
\eta_j(t)\nabla_{\theta_j^i} & = \eta_j(t)\frac{\partial L}{\partial \theta_j^i} = \eta_j(t)\frac{\partial L}{\partial g(\theta_j^i)}\frac{\partial g(\theta_j^i)}{\partial \theta_j^i} \\
& = \eta_j(t)\frac{\partial g(\theta_j^i)}{\partial \theta_j^i}\Big(\alpha\frac{\partial L_{computational}}{\partial g(\theta_j^i)} + \frac{\partial L_{original}}{\partial g(\theta_j^i)}\Big) \\ 
& = \frac{\gamma \eta(t)}{\widehat{\lambda_j}(t)}\frac{\partial g(\theta_j^i)}{\partial \theta_j^i}\Big(\alpha\widehat{\lambda_j}(t)\frac{\partial c_j(t)}{\partial g(\theta_j^i)} + \frac{\partial L_{original}}{\partial g(\theta_j^i)}\Big) \\ 
& =  \gamma\eta(t)\frac{\partial g(\theta_j^i)}{\partial \theta_j^i}\big(\alpha + \frac{1}{\widehat{\lambda_j}(t)}\frac{\partial L_{original}}{\partial g(\theta_j^i)}\big)\; ,
\end{split}
\end{equation}
where we substituted, respectively, Eqn.~\ref{eq:total_loss} in the second line, Eqn.~\ref{eq:computation_loss} and Eqn.~\ref{eq:gradient_adjustment} in the third line, and Eqn.~\ref{eq:channel_sum} in the last line.
The role of parameter $\gamma$ is to balance out the large values of the term $\frac{1}{\widehat{\lambda}(t)}$, since $\widehat{\lambda}(t)$ is generally small. 

In Eqn.~\ref{eq:gradient_adjustment_derivation} we observe the two components of the gradient produced by the computational loss and the original loss. The first component attempts to gate the channel, while the other retains as much relevant information as possible. The gradient adjustment from Eqn.~\ref{eq:gradient_adjustment} is applied on both parts, which is crucial for a balanced optimization. The last line shows how the gradient adjustment cancels out $\widehat{\lambda_j}(t)$ for the pruning cost component, making the impact of the pruning loss function on the gradient update similar for all edges.

Another insight from Eqn.~\ref{eq:gradient_adjustment_derivation} is the \textit{self-pruning} phenomenon we have occasionally observed, where the original loss component contributes to the removal of certain channels on its own (even with a zero pruning loss). This could occur if a channel has a negative impact on the original loss, which in turn, produces a larger update on the gating weight compared to the update on the channel weight; thus, the channel is pruned faster then it can adjust. To alleviate this issue, we calibrate $\gamma$ by performing training with no pruning loss, and setting it to balance reasonable conversion time with minimal self-pruning.

\section{Experiments}

\subsection{Methodology and Setup}
\label{Methodology and setup}

The lack of a comparison standard for pruning results is discussed in~\cite{blalock2020state}, listing different types of pruning, different benchmark types, and a plethora of different network architectures on which various pruning methods are benchmarked. Hence, comparing between pruning methods is only relevant for compatible criteria, i.e., same pruning type (e.g., structured pruning), optimized parameter (e.g., FLOPs reduction), and architecture. (From a practical consideration, we only compared with structured channel pruning methods.)
Similarly to most other works, we performed pruning on the ImageNet ILSVRC2012~\cite{russakovsky2015imagenet} and CIFAR-10~\cite{krizhevsky2009learning} classification challenges. However, we believe that results for CIFAR-10 are not very indicative, due to the small dataset, the large variation in accuracy~\cite{frankle2018lottery}, and the fact that the dataset does not adequately represent ``real-world'' data.
We chose the ResNet architecture, which enabled demonstrating Gator's pruning capability for a non-sequential architecture and comparing it to other works. Note that newer and smaller architectures would require a much greater effort in recreating proper training and pruned versions, not to mention that there are very few results to compare to.

In both experiments, we trained our model with the standard SGD optimizer, setting the momentum to $0.9$ and the weight decay to $0.0001$. We initialized the gating probability to $0.005$,
which was found empirically to have a low impact on the network's initial accuracy, while still enabling the pruning in a relatively low number of epochs. Note that a lower initial gating probability would have required more steps to reach the pruning threshold probability set to $0.5$. Crossing the threshold probability marks a channel as permanently pruned.
We performed iterative pruning with an increasing global weight $\alpha$. Each pruning iteration consisted of training with gating, and of additional fine-tuning without gating. The next iteration loaded the weights produced by the gating phase.

\subsection{Results for ResNet-50 on ImageNet}

\begin{figure}[!tbp]
\subfigure[]{\includegraphics[width=6cm]{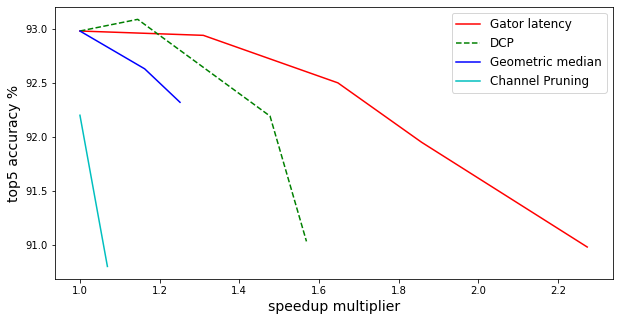}}
\hfill
\subfigure[]{\includegraphics[width=6cm]{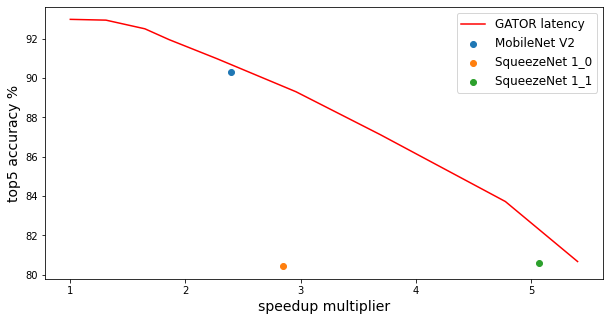}}
\caption{ImageNet (ILSVRC-12) top-5 accuracy (\%) as a function of latency speedup for ResNet-50 pruning by Gator compared with (a) other pruning networks and (b) smaller networks for larger scale pruning.}
\label{fig:resnet50_comparison_other_works}
\vskip -0.15in
\end{figure}

\begin{table*}[t]
\caption{Comparison of pruned ResNet-50 on ImageNet for Gator optimized on FLOPs and GPU latency; number of Gator variant indicates pruning weight $\alpha$; results are ordered and divided to categories by FLOPs reduction ($\%$)}
\label{tbl:imagenet_resnet50}
\vskip 0.1in
\begin{center}
\begin{small}
\begin{sc}
 \begin{threeparttable}
\begin{tabular}{lccccc}
\toprule
\begin{tabular}{@{}c@{}}Pruning \\Algorithm\end{tabular} & \begin{tabular}{@{}c@{}c@{}} Top-1 \\acc.\% \end{tabular} & \begin{tabular}{@{}c@{}c@{}} Top-5 \\acc.\% \end{tabular} & \begin{tabular}{@{}c@{}} FLOPs \%\\reduction\end{tabular} & \begin{tabular}{@{}c@{}} Memory \%\\ reduction\end{tabular} & \begin{tabular}{@{}c@{}} Practical \\Speed-up \tnote{1}\end{tabular} \\
\midrule
channel pruning \cite{he2017channel} & 72.30 & 90.80 & 24.14 & 15.52 & 1.069 \\
\textbf{Gator latency 0.25} & 76.00 & 92.94 & 26.75 & 5.22 & \textbf{1.310} \\
\textbf{Gator flops 0.25} & 75.89 & 92.93 & 31.48 & 8.88 & 1.200 \\
DCP 30 \cite{zhuang2018discrimination} & \textbf{76.35}\tnote{2} & \textbf{93.09} \tnote{2} & 36.04 & 33.45 & 1.145 \\
Geometric 30\% \cite{he2019filter} & 75.59 & 92.63 & 41.89 & 37.64 & 1.162 \\
\hline
\textbf{Gator latency} 0.5 & \textbf{75.28} & 92.50 & 48.27 & 18.05 & \textbf{1.647} \\
\textbf{Gator flops 0.5} & 75.19 & \textbf{92.61} & 49.91 & 18.96 & 1.442 \\
Geometric 40\% & 74.83 & 92.32 & 53.16 & 47.72 & 1.251 \\
DCP 50 & 74.88 & 92.19 & 55.77 & 51.63 & 1.477 \\
PCAS \cite{yamamoto2018pcas} & 72.68 & 91.09 & 56.70 & 51.20 & -- \\
\hline
\textbf{Gator latency 1} & \textbf{74.24} & 91.95 & 61.17 & 31.18 & \textbf{1.857} \\
\textbf{Gator flops 1} & 74.14 & \textbf{91.99} & 62.64 & 31.40 & 1.612 \\
DCP 70 & 72.65 & 91.03 & 71.10 & 66.00 & 1.569 \\
\hline
\textbf{Gator flops} 2 & 72.36 & 90.97 & 76.57 & 55.15 & 2.106 \\
\bottomrule
\end{tabular}
  \begin{tablenotes}
  \item[1] \footnotesize Speedup multiplier for batch 32 on single GTX 1080 Ti GPU
  \item[2] \footnotesize Better accuracy then baseline ResNet-50.
  \end{tablenotes}
   \end{threeparttable}
\end{sc}
\end{small}
\end{center}
\vskip -0.1in
\end{table*}

\begin{table}[!tbp]
\caption{Comparison of small networks to high rate pruned ResNet-50 on ImageNet for Gator optimized on GPU latency; number of Gator variant indicates pruning weight $\alpha$; results are ordered and divided to categories by speedup}
\label{tbl:imagenet_resnet50_speedup}

\begin{center}
\begin{small}
\begin{sc}
 \begin{threeparttable}
\begin{tabular}{lcccc}
\toprule
\begin{tabular}{@{}c@{}}Pruning \\Algorithm\end{tabular} & \begin{tabular}{@{}c@{}c@{}} Top-1 \\acc.\% \end{tabular} & \begin{tabular}{@{}c@{}c@{}} Top-5 \\acc.\% \end{tabular} & FLOPs &  \begin{tabular}{@{}c@{}} Speed \\ up \end{tabular} \\
\midrule
\textbf{Gator latency 2} & \textbf{72.69} & \textbf{90.98} & 973M & 2.273 \\
Mobilenet V2 & 71.87 & 90.29 & 301M & \textbf{2.399} \\
\hline
squeezenet 1.0 & 58.10 & 80.42 & 888M & 2.843 \\
\textbf{Gator latency 4} & \textbf{69.48} & \textbf{89.3} & 551M & \textbf{2.959} \\
\hline
\textbf{Gator latency 16} & \textbf{61.07} & \textbf{83.72} & 191M & 4.777 \\
squeezenet 1.1
& 58.19 & 80.62 & 387M & 5.070 \\
\textbf{Gator latency 20} & 56.89 & 80.67 & 134M & \textbf{5.404} \\

\bottomrule
\end{tabular}
\end{threeparttable}
\end{sc}
\end{small}
\end{center}
\vskip -0.3in
\end{table}

For pruning ResNet-50 trained on ImageNet, we used the pretrained weights.
The batch size was 256, which was split between two GPUs. However, due to the lack of RAM, we used gradient accumulation and split each half batch into 4 parts, training equivalenty on 8 GPUs with sub-batches of size 32 each. (This only affects the batch normalization.) We trained with gates for $30$ epochs, using learning rates of 0.01 and 0.001 for the first 20 epochs and the latter 10 epochs, respectively. Table~\ref{tbl:imagenet_resnet50} contains fine-tuning results for lower pruning rates ($\alpha \leq 2$). We trained, in this case, for $40$ epochs with a learning rate step policy of 0.001, 0.0001, and 0.00001 for 20,10,and 10 epochs, respectively. For high pruning rates, we trained for 20 epochs (given the available resources under the time constraints), with a learning rate step policy of 0.001, 0.0001, and 0.00001 for 10,5,and 5 epochs, respectively (see Table~\ref{tbl:imagenet_resnet50_speedup}).
For image preprocessing, we applied mean subtraction and standard deviation normalization for each color channel. Data augmentation consisted of selecting a random 
$224 \times 224$ crop from the original image and applying to it a random horizontal flip. Practical speedup was measured by timing an inference run in Pytorch on GTX 1080 Ti GPU with a batch of size 32 for $224 \times 224$ images. 

Table~\ref{tbl:imagenet_resnet50} contains comparative results of optimized Gator pruning on FLOPs and latency, with other methods (also illustrated in Fig. \ref{fig:resnet50_comparison_other_works}(a)). Note the very similar accuracy obtained by the two optimization methods and how each type is better suited for its designated reduction metric. 
(Although there is a single instance that outperforms Gator (optimized on FLOPs reduction) for lower pruning rates, note that the accuracy reported in this
case~\cite{zhuang2018discrimination} exceeds the original ResNet-50 accuracy; this could be explained by the use of auxiliary classification losses, known to enhance the performance~\cite{szegedy2016rethinking}.
Thus, we believe this isolated instance is not a fair comparison.)

Regarding the latency speedup, Gator outperforms all other methods, achieving a speedup of 1.4 for some pruning rates.

Results for higher pruning rates are available in Table \ref{tbl:imagenet_resnet50_speedup}. We compared Gator's latency optimization to that of much smaller architectures, such as SqueezeNet \cite{iandola2016squeezenet} and MobileNet V2 \cite{sandler2018mobilenetv2}, a newer architecture containing speed-efficient elements, such as inverse bottlenecks and grouped convolutions. According to Fig.~\ref{fig:resnet50_comparison_other_works}(b), our method slightly outperforms both of these small architectures.

\subsection{Results for ResNet-56 on CIFAR-10}

We trained ResNet-56 on CIFAR-10 from scratch for $240$ epochs with a batch size of $128$. The learning rate was $0.1$ for the first $120$ epochs, and was changed to 0.01 and 0.001 at epochs 120 and 180, respectively.
For pruning, we trained with gating for $90$ epochs, starting with at a learning rate of $0.01$ and decreasing it to $0.001$ at epoch $60$. This was followed by fine-tuning for another $30$ epochs at the same learning rate. 
Image preprocessing consisted of mean subtraction and standard deviation division for each channel. Data augmentation consisted of random cropping and random horizontal flipping. The results are available in Table \ref{tbl:cifar10_resnet-56}, where we see a significant reduction of FLOPs with little accuracy loss for Gator which also and best performance on memory reduction compared to other models. In addition Gator was able to prune entire ResNet blocks, thereby reducing the number of layers.

\begin{table}[!tbp]
\caption{Pruning results of ResNet-56 on CIFAR-10 using Gator optimization on FLOPs and memory vs. other methods; baseline network accuracy is listed for comparison as it differs across methods}
\label{tbl:cifar10_resnet-56}
\begin{center}
\begin{small}
\begin{sc}
 \begin{threeparttable}
\begin{tabular}{lcccc}
\toprule
\begin{tabular}{@{}c@{}}Pruning \\Algorithm\end{tabular} & \begin{tabular} {@{}c@{}} Baseline \\ acc. \% \end{tabular} & \begin{tabular} {@{}c@{}} Acc. \\ \% \end{tabular} & \begin{tabular}{c@{}c@{}} FLOPs \% \\ reduction \end{tabular} & \begin{tabular}{c@{}c@{}} Memory \% \\ reduction \end{tabular}\\ 
\midrule

\textbf{Gator flops 1} & 93.63 & 93.58 & 43.39 & 33.50 \\
\textbf{Gator flops 2} & 93.63 & 93.14 & 54.62 & 44.39 \\
\textbf{Gator memory 0.25} & 93.63 & 93.66 & 33.21 & 42.70 \\
\textbf{Gator memory 0.5} & 93.63 & 92.59 & 57.98 & \textbf{75.08} \\

DCP & 93.80 & 93.49 & 49.75 & 49.24 \\
DCP Adapt & 93.80 & \textbf{93.81} & 47.09 & 70.33 \\
Geometric 40 & 93.59 & 93.49 & 52.30 & -- \\
Channel Pruning & 92.80 & 91.80 & 50.00 & -- \\
PCAS & 93.04 & 93.58 & \textbf{54.80} & 53.70 \\
\hline
\textbf{Gator flops 16} &  93.63 & 91.50 & \textbf{79.33} & 73.40 \\
\hline
\textbf{Gator memory 2} &  93.63 & 90.99 & 76.05 & \textbf{90.87} \\

\bottomrule
\end{tabular}
\end{threeparttable}
\end{sc}
\end{small}
\end{center}
\vskip -0.3in
\end{table}

\section{Conclusions}

In this paper we presented Gator, a novel structural channel-pruning method, which utilizes dynamic channel gating during training to obtain smaller effective networks. 
We showed how to prune non-sequential architectures by using a hyper-graph formulation to encapsulate the mapping of dependencies between input and output channels of different convolutions.
We also demonstrated how Gator can accommodate a variety of auxiliary cost functions to optimize FLOPs, memory, and latency reduction due to pruning.
Latency reduction, in particular, is important for allowing to produce very light networks designed for small devices. 
Finally, we showcased SOTA results of pruning ResNet-50, trained on ImageNet, with respect to FLOPs reduction and practical speedup on actual hardware.

\bibliography{gator_bib}
\bibliographystyle{splncs04}

\end{document}